\documentclass[sigconf,anonymous=False]{acmart}

\usepackage{xspace}
\usepackage{balance}
\usepackage{enumitem}
\usepackage{ulem}
\usepackage[ruled,lined,linesnumbered,noend]{algorithm2e}
\usepackage{cleveref}
\usepackage{amsmath}
\usepackage{arydshln}
\usepackage{booktabs}
\usepackage{multirow}
\usepackage{afterpage}
\usepackage{lipsum}
\usepackage{fancyhdr}

\def\revisionmode{1}

\ifdefined\revisionmode
\newcommand{\yiwei}[1]{{\color{orange} {\bf Yiwei:} #1}}
\newcommand{\ziyun}[1]{{\color{red} {\bf Ziyun:} #1}}
\newcommand{\phil}[1]{{\color{blue} {\bf Phil:} #1}}
\newcommand{\sai}[1]{{\color{olive} {\bf Sai:} #1}}
\newcommand{\syed}[1]{{\color{purple} {\bf Syed:} #1}}
 \newcommand{\kleber}[1]{{\color{magenta} {\bf Kleber:} #1}}
\newcommand{\barbara}[1]{{\color{green} {\bf Barbara:} #1}}
\else
\newcommand{\yiwei}[1]{}
\newcommand{\ziyun}[1]{}
\newcommand{\phil}[1]{}
\newcommand{\sai}[1]{}
\newcommand{\syed}[1]{}
\newcommand{\kleber}[1]{}
\newcommand{\barbara}[1]{}
\fi

\definecolor{forestgreen}{rgb}{0.13, 0.55, 0.13}

\renewcommand{\emph}{\textit}

\ifdefined\confversion
\ifdefined\nocolor

\else

\fi
\else

\fi

\newcommand{\sota}{state-of-the-art\xspace}

\newcommand{\myparagraph}[1]{\smallskip\noindent {\bf #1.}}

\newcommand{\hide}[1]{}

\newcommand\blfootnote[1]{%
  \begingroup
  \renewcommand\thefootnote{}\footnote{#1}%
  \addtocounter{footnote}{-1}%
  \endgroup
}

\AtBeginDocument{%
  \providecommand\BibTeX{{%
    \normalfont B\kern-0.5em{\scshape i\kern-0.25em b}\kern-0.8em\TeX}}}

\acmConference[DAC]{Design Automation Conference}{June 2024}{San Francisco, CA, USA}

\setcopyright{none}
\renewcommand\footnotetextcopyrightpermission[1]{}

\begin{document}

\pagestyle{empty}

\title{Neural Architecture Search of Hybrid Models for NPU-CIM Heterogeneous AR/VR Devices}

\author[Y. Zhao, et al.]
{Yiwei Zhao$^1$,
Ziyun Li$^2$,
Win-San Khwa$^3$,
Xiaoyu Sun$^3$,
Sai Qian Zhang$^4$,
Syed Shakib Sarwar$^2$,
Kleber Hugo Stangherlin$^2$,
Yi-Lun Lu$^3$,
Jorge Tomas Gomez$^2$,
Jae-Sun Seo$^2$,
Phillip B. Gibbons$^1$,
Barbara De Salvo$^2$,
Chiao Liu$^2$}
\affiliation{\institution
{$^1$Carnegie Mellon University,
$^2$Meta Reality Labs Research,
$^3$TSMC Corporate Research,
$^4$New York University}
\country{}
}
\email{yiweiz3@andrew.cmu.edu}
\email{liziyun@meta.com}

\renewcommand{\shortauthors}{Yiwei Zhao, et al.}

\renewcommand\abstractname{\textsc{ABSTRACT}}
\begin{abstract}
Low-Latency and Low-Power Edge AI is essential for Virtual Reality and Augmented Reality applications.
Recent advances show that hybrid models, combining convolution layers (CNN) and transformers (ViT), often achieve superior accuracy/performance tradeoff on various computer vision and machine learning (ML) tasks.
However, hybrid ML models can pose system challenges for latency and energy-efficiency due to their diverse nature in dataflow and memory access patterns.
In this work, we leverage the architecture heterogeneity from Neural Processing Units (NPU) and Compute-In-Memory (CIM) and perform diverse execution schemas to efficiently execute these hybrid models.
We also introduce H4H-NAS, a Neural Architecture Search framework to design efficient hybrid CNN/ViT models for heterogeneous edge systems with both NPU and CIM.
Our H4H-NAS approach is powered by a performance estimator built with NPU performance results measured on real silicon, and CIM performance based on industry IPs.
H4H-NAS searches hybrid CNN/ViT models with fine granularity and achieves significant (up to $1.34\%$) top-1 accuracy improvement on ImageNet dataset.
Moreover, results from our Algo/HW co-design reveal up to $56.08\%$ overall latency and $41.72\%$ energy improvements by introducing such heterogeneous computing over baseline solutions.
The framework guides the design of hybrid network architectures and system architectures of NPU+CIM heterogeneous systems.
\blfootnote{This is an extended version of the work titled ``H4H: Hybrid Convolution-Transformer Architecture Search for NPU-CIM Heterogeneous Systems for AR/VR Applications'' presented in the 61th Design Automation Conference (DAC '24).}
\end{abstract}


\renewcommand\keywordsname{\textsc{KEYWORDS}}
\keywords{neural architecture search, compute-in-memory, neural processing unit, near-memory compute}

\maketitle

\section{Introduction}

Virtual Reality (VR) and Augmented Reality (AR) are increasingly prevailing as key next-generation human-oriented computing platforms~\cite{abrash2021creating}.
The recent advances in artificial intelligence (AI) power multiple applications in AR/VR to revolutionize how people communicate with each other, improving people's productivity and how people interact with the digital world.
These applications typically involve running multiple Deep Neural Network (DNN) inferences for different tasks such as hand tracking~\cite{han2020megatrack}, eye tracking~\cite{plopski2022eye}, object detection~\cite{ghasemi2022deep}, photo realistic avatars~\cite{xiang2022dressing}, etc.

Typically, in order to meet the low latency requirements of these AR/VR applications (such as hand tracking and detection) and to preserve user privacy, most DNN inferences need to be processed locally on AR/VR devices.
In addition, given the limited on-device (on-AR/VR glass) compute, memory capacity, and power budget, as well as the recent emergence of smart cameras ~\cite{liu2020intelligent, imx}, the on-device processing is heavily distributed between the main SoC and multiple intelligent sensors, making part of processing reside locally on intelligent sensors ~\cite{gomez2022distributed,dong2022splitnets}.

These intelligent sensors, although limited in compute / memory capacity due to area constraints, are required to achieve high energy efficiency for ML tasks with ultra-low latency.
Meanwhile, DNN models for these applications are becoming increasingly diverse to improve task performance, even if they are targeting similar classes of workloads.
For instance, in computer vision (CV), ResNet~\cite{he2016deep}, MobileNet-v2~\cite{sandler2018mobilenetv2} and vision transformers (ViT)~\cite{dosovitskiy2020image,liu2021swin} have completely different basic block structures and require varied execution schemas.
This poses difficulty in designing general-purpose accelerators that are efficient on all these various models:
An accelerator heavily optimized for one generation of model is often less efficient when new models are invented.

Neural Processing Units (NPUs) have emerged as a promising means for addressing these
challenges and meeting the stringent energy/latency requirement for edge AI,
with the technology recently maturing into widespread adoption in commercial products~\cite{u55,samsungnpu}.
Many \sota NPUs adopt systolic array 
structures and are highly-efficient in compute-intensive workloads.
For example, ARM Ethos-U55~\cite{u55} and U65~\cite{u65} are highly-efficient on convolution layers (CNN).

As the compute capacity increases with the rise of NPUs, however, the frequent data movement between memory and processor dominates energy and latency costs.
To address this, compute-in-memory (CIM) has re-emerged as an effective architecture for reducing data movement.
In CIM, computing elements are close to (near-memory computing (NMC)~\cite{MRAM1,mutlu2023primer,kang2022pimtreevldb}) or even merged with (in-memory computing (IMC)~\cite{RRAM1,RRAM3,PCM1}) memory, enhancing latency and energy efficiency.

CIM-based ML tasks rely on dense on-chip storage of model weights in order to avoid energy costs associated with reading from external memory.
For instance, there are highly-efficient CIM accelerators optimized for MobileNetv2~\cite{wang202328nm} and transformers~\cite{tu2023multcim,liu202328nm}, which leverage their efficient local data processing for memory-bounded workloads.
The dense storage also makes CIM based on \emph{non-volatile memory} (e.g., resistive RAM (ReRAM)~\cite{RRAM1,RRAM3}, phase-change RAM (PCRAM)~\cite{PCM1}, magnetic RAM (MRAM)~\cite{MRAM1,MRAM2}) an attractive alternative to
SRAM/DRAM CIM~\cite{SRAM1,dram1}, by mitigating the leakage and refresh costs of volatile memories and avoiding reload latency on initialization, especially in "mostly-off" settings.

In this work, we propose a generic design that combines both NPUs and CIMs, leveraging the architectural heterogeneity from NPUs and CIMs to accelerate AI edge systems with diverse dataflows arising from our hybrid CNN/ViT models.
We also introduce a neural architecture search (NAS) framework to co-design hybrid CNN/ViT models to achieve the best accuracy/performance trade-offs for heterogeneous architectures. 

Our key contributions and novel aspects are as follows:

\begin{itemize}
    \item We present \underline{H4H}-NAS: A neural architecture search framework to design and search efficient
    \underline{H}ybrid CNN/ViT models \underline{for} usage on
    \underline{H}eterogeneous edge compute featuring NPU and CIM.
    \item We build a workflow with a system modeling tool using post-silicon results for NPU and industry IP-based results for CIM to guide the efficient model development process.
    \item We propose system-level improvements on current CIM-based designs, including adding multiple compute-units in CIMs and multiple macros in the system, to further improve system performance on ML workloads.
\end{itemize}

\section{Preliminaries}

In this section, we provide preliminaries and the motivation of our methods.
We first evaluate the characteristics of NPU and CIM, showing they are very efficient at compute-intensive and memory-intensive workloads, respectively.
Then we discuss the recent advances of hybrid models from academia and industry for AI edge systems.
This background motivates our Algorithm/Hardware co-design of efficient hybrid models and accelerating them with a heterogeneous system in AI edge devices.

\subsection{Heterogeneous Platforms w/ NPU and CIM}
\label{subsec:preliminary_heterogeneous}

To architect AI edge systems with both NPU and CIM, we first collect and analyze performance data from real-world silicons of NPUs and SPICE-simulated industry CIM IPs.
These collected data points provide an accurate modeling of the energy and performance of a heterogenouse system for our framework.

\begin{figure}[t]
    \centering
    \includegraphics[width=0.55\linewidth]{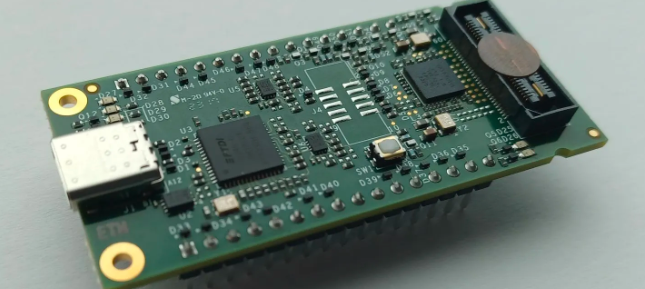}
    \vspace{-2em}
    \caption{Silicon for testing ARM Ethos-U55 NPU.\vspace{-2em}}
\label{fig:evb}
\end{figure}

\myparagraph{NPU}
We use the ARM Ethos-U55~\cite{u55} as a typical example of an NPU on edge devices.
Our test silicon, shown in \Cref{fig:evb}, is fabricated and measured using 7nm FinFET technology. 

We test different DNN models and layers---regular/depth-wise/point-wise convolutions and fully-connected layers---using the NPU in the chip with the ARM ethos-u-vela toolchain.
All experiments are performed with a batch size of $1$, which is common in edge inference applications.
System metrics measured are execution latency and energy consumption.

\begin{figure*}[t]
    \centering
    \includegraphics[width=\linewidth]{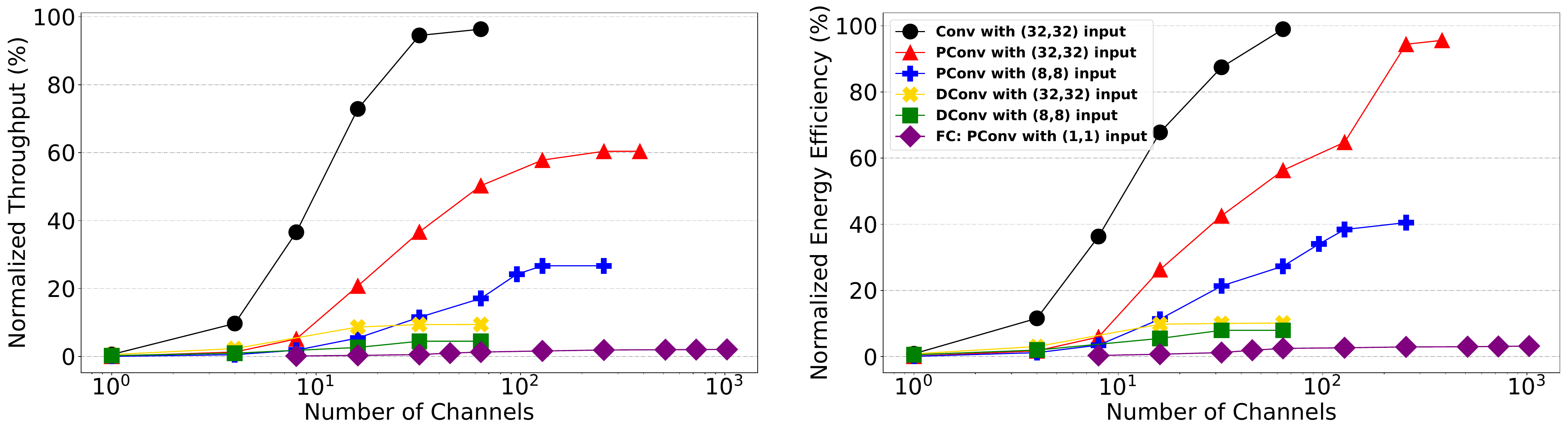}
    \vspace{-2.2em}
    \caption{Throughput and energy efficiency of Ethos-U55 NPU execution of different layers, normalized by the theoretical best performance on U55. Conv and Dconv respectively stands for regular convolution and depthwise convolution with (3,3)-kernels. PConv represents pointwise convolution with (1,1)-kernel. FC refers to fully-connected layers.\vspace{-0.5em}}
\label{fig:u55}
\end{figure*}

\Cref{fig:u55} shows the throughput and energy efficiency of typical layers executed on U55, both of which are normalized by the theoretical performance of the NPU.
In summary, different layer types illustrate different execution efficiencies, but they all follow a trend of ``increasing then saturating performance'' as data sizes grow larger.

\begin{figure}[t]
    \centering
    \includegraphics[width=\linewidth]{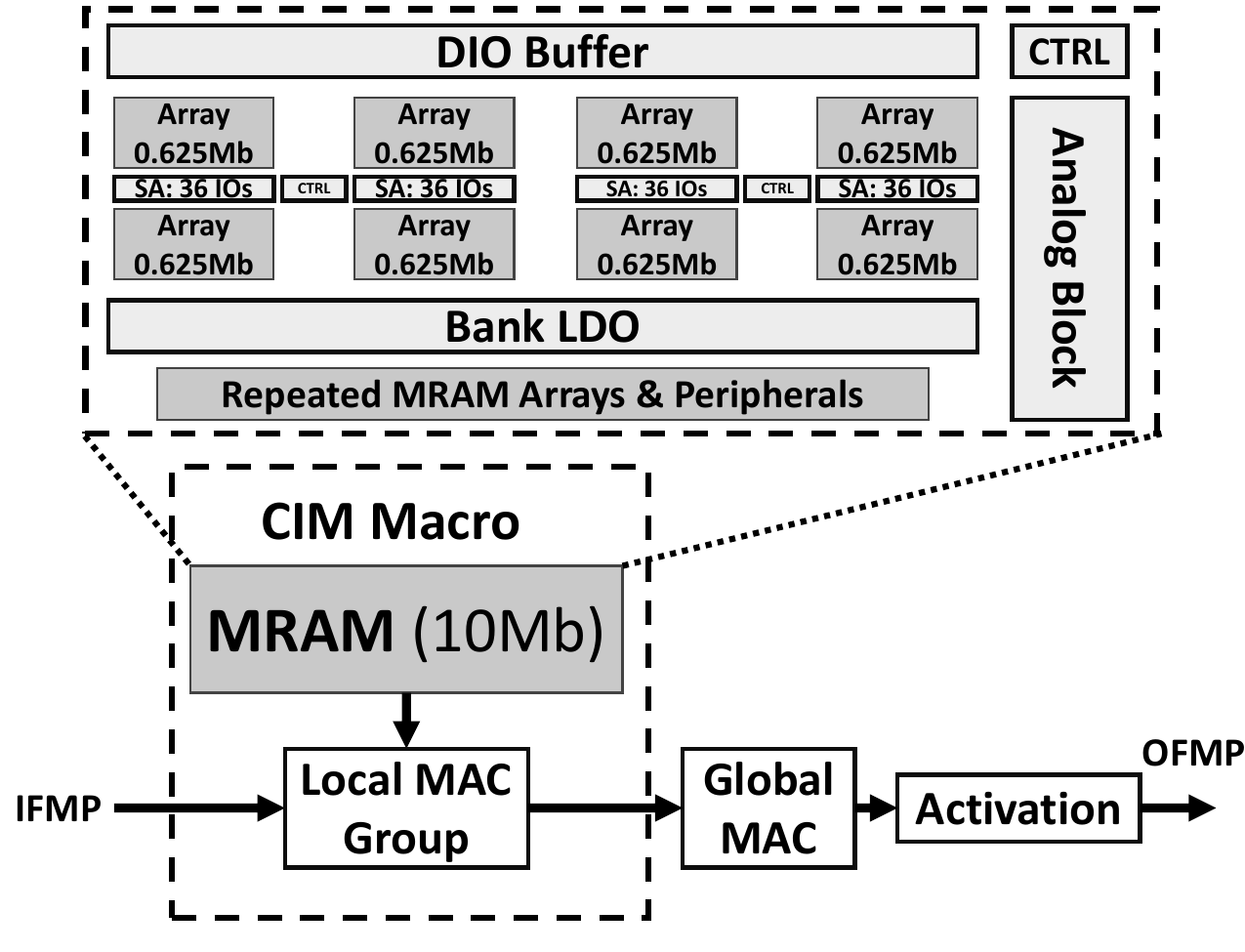}
    \vspace{-3em}
    \caption{Architecture layout of the MRAM CIM macro. I/OFMP stand for input/output feature maps.\vspace{-1.5em}}
\label{fig:cim_layout}
\end{figure}

\myparagraph{CIM}
We acquire our CIM data on a digital-based NMC MRAM-based CIM macro.
The non-volatility of MRAM helps reduce wake-up overhead on edge AR/VR applications.
The MRAM macro is evaluated in 7nm technology (projected from 16nm designs) for fair comparison with NPU.
It is implemented based on production designs~\cite{DTP1,DTP2} with read optimization for lower read energy.
Each MRAM macro has 10Mb memory capacity and can compute the 16 accumulations of 9 products between 8-bit input and 8-bit weight.
The memory and the computation peripheral occupy approximately 0.9mm$^2$ and 0.15mm$^2$, respectively.
\Cref{fig:cim_layout} shows the overall architecture of our MRAM CIM macro.

We focus on performance of CIM executing memory-bounded layers, such as depthwise convolutions and fully-connected layers, as NPU performs suboptimally on these workloads.
We also acquire CIM performance on pointwise convolution, as there is potential in leveraging this workload over NPU results.

\begin{figure}[t]
    \centering
    \includegraphics[width=0.9\linewidth]{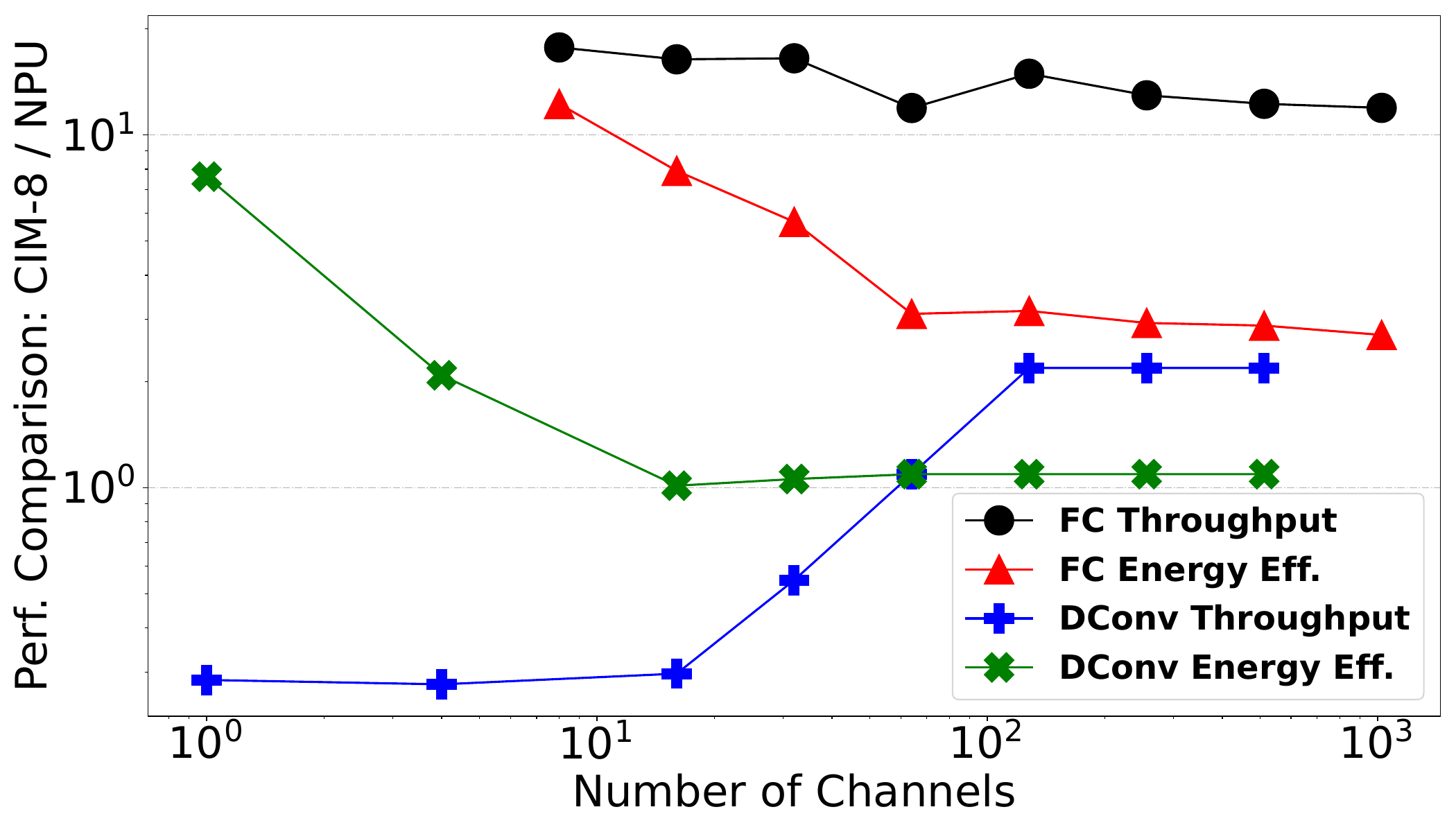}
    \vspace{-1.2em}
    \caption{The comparative ratio of throughput and energy efficiency between a system with 8 CIM macros and a U55-only system when executing fully-connected layers and depthwise convolution with (3,3)-kernel and (32,32)-input.\vspace{-1.5em}}
\label{fig:cim_layer_wise}
\end{figure}

\Cref{fig:cim_layer_wise} shows the comparative ratio of throughput and energy efficiency between eight MRAM CIM macros and one U55 NPU.
It is indicated that a system with multiple CIM macros working together can potentially outperform NPU on memory-bounded DNN layers in both throughput and energy efficiency, if given practical layer configurations from existing ML models.

\subsection{Hybrid Models with CNN and ViT}
\label{subsec:preliminary_model}

With recent developments of ML, hybrid models become increasingly prevailing due to their potentially better accuracy in CV than traditional networks such as VGG/ResNet. 
Two common examples are the introduction of depthwise convolution and transformers.

\myparagraph{IRB}
Inverted residue bottleneck block (IRB) from MobileNet-v2~\cite{sandler2018mobilenetv2} is widely used for low-latency and efficient edge AI inferences.
An IRB block consists of two pointwise and one depthwise convolutions and enables significantly reduced compute and memory.

\myparagraph{ViT}
Vision transformers (ViT)~\cite{dosovitskiy2020image,liu2021swin} recently emerges and adopts transformers as basic blocks.
A transformer block is composed of Q/K/V generators, head-level multiplication, layer normalization, softmax, positional encoding and multilayer perceptrons (MLP).

\myparagraph{Hybrid Networks}
In addition to diversity resulting from different components inside one block, some recent models use multiple types of blocks in their networks.
For instance, SAM ~\cite{kirillov2023segment}, LeViT ~\cite{graham2021levit} and AlterNet ~\cite{park2021vision} combine both CNNs and ViTs in their models to achieve state-of-the-art performance.

\subsection{Discussion}
Although hybrid models can be more accurate (\Cref{subsec:preliminary_model}), their different layers typically require HW support for diverse executing schemas in inference and training.
Different layers may also trigger different requirements in compute and memory access patterns, putting different pressure on compute units and memory devices.

Meanwhile, results in \Cref{subsec:preliminary_heterogeneous} indicate that the heterogeneity between NPU and CIM could potentially offer solutions for efficient execution of hybrid models, given NPU achieves high efficiency on compute-intensive workloads while CIMs are very efficient when computation is bottlenecked by memory accesses.

In this work, we focus on the model search and system improvements for heterogeneous systems consisting of both NPU and CIM, as a potential solution to the mentioned design difficulties.
We select ARM ethos-u55 NPU and digital-based NMC MRAM CIM as representatives of the hardware components, as we think they are adaptive to multiple workloads,
but our methods can be used in other hardware designs, such as analog-based IMC CIMs.

\section{Methodology}

\myparagraph{Workflow Overview}
We build a workflow to co-design the algorithm/hardware for efficient inference with hybrid CNN/ViT models for heterogeneous edge systems featuring NPU and CIM.
The workflow targets CV tasks in AR/VR applications and incorporates real-world resource constraints for an AI edge system such as intelligent cameras.
We leverage the two stage Neural Architecture Search (NAS) to automate the process and efficiently search the optimized models.
We aim to answer two fundamental questions by analyzing the discovered models:
(1) To ML researchers: What model architecture is preferred by heterogeneous edge devices?
(2) To system designers: How to efficiently map and process hybrid models using NPU and CIM?

\subsection{NAS for Hybrid Models}
Previous NAS approaches that use evolutionary search ~\cite{real2017large} or reinforcement learning ~\cite{zoph2016neural} require excessive training due to large numbers of models trained in a single experiment.
Recent NAS advances decouple model training and architecture search into two separate stages ~\cite{cai2019once} that significantly reduces training cost.
We build our H4H-NAS based on this two-stage NAS framework that is broadly applied in AR/VR.
In the first stage, we train a model that optimizes a \textit{supernet} and all its sub-networks with weight-sharing.
In the second stage, \textit{subnets} are searched based on system constraints to yield the best task and system performance trade-off.

Although prior works deploy efficient NAS for edge model designs for CNN ~\cite{cai2019once} and ViT ~\cite{gong2022nasvit}, they only provide results on pre-defined supernet structures with minimal flexibility.
For example, NASViT~\cite{gong2022nasvit} restricted transformer layers to be used only at the end of the network, instead of fusing transformers into the search space flexibly.
Previous NAS approaches also focuses on FLOP and parameter optimizations without taking NPU/CIM heterogenous architectures into consideration.

In this work, we adopt two-stage NAS as the core strategy in our algorithm-system co-design and focus on enabling a flexible search space of hybrid models and deploying on heterogeneous architectures built from industrial IPs.

\myparagraph{Search Space}
We summarize the detailed search dimensions of our search space in~\Cref{tab:searchs_space}.
For inverted residual bottleneck blocks (IRB), we search for the number of output channels (width), the number of layers in a single block (depth), and the expansion ratio of depthwise convolutions.
Stride=2 only applies to the first layer in each block.
For vision transformer encoders ~\cite{dosovitskiy2020image}, we search for the Q/K/V dimension (width), the number of layers in a single block (depth), and the expansion ratio of MLP.
We fix the number of input channels and output channels of a transformer block to be equal, to enable unchanged residues to pass over transformer blocks.
We use (3,3)-sized kernels in all convolution layers and 8-dimension heads in all transformers. 

\begin{figure}[t]
    \centering
    \includegraphics[width=\linewidth]{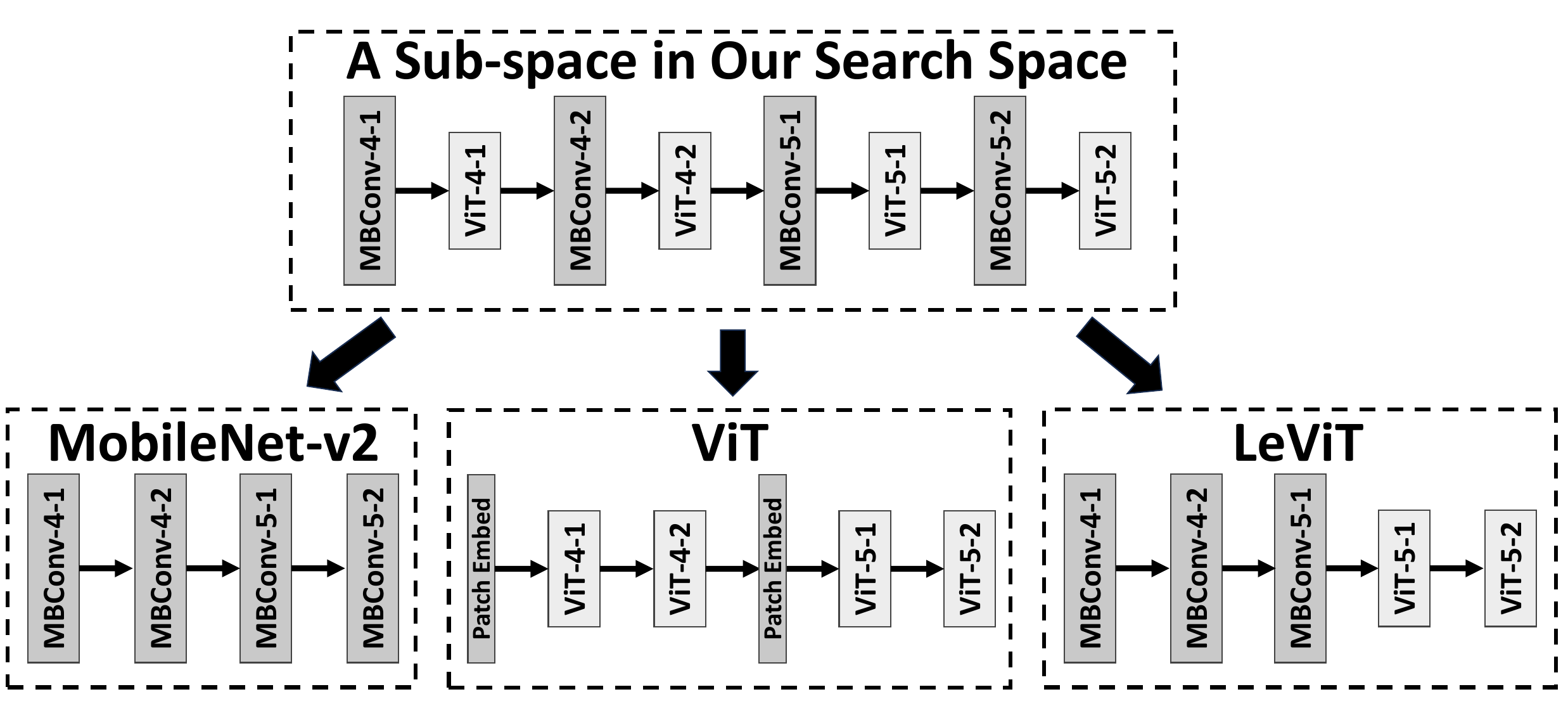}
    \vspace{-2.2em}
    \caption{An example of how our search space can be flexibly reduced to basic blocks of different existing model types.\vspace{-1.5em}}
\label{fig:nas_search_block}
\end{figure}

We build our supernet structure using repeated ``convolution + transformer'' blocks, as suggested in~\cite{park2021vision}.
Note that our supernet can be flexibly reduced to either an IRB-only model, a ViT-only model, or a ``first CNN then pure ViT'' structure similar to LeViT~\cite{graham2021levit}, as shown in \Cref{fig:nas_search_block}.
This ensures the superior flexibility in the supernet design.
This supernet can eventually be turned into diverse hybrid models during the subnet search.

\begin{table}
\centering
    \begin{tabular}{|c|cccc|}
    \hline
    Block & Width & Depth & Exp. Ratio & Stride \\
    \hline
    Conv-0 & $16\sim32$ & $1$ & - & $2$ \\
    \hdashline[2pt/5pt]
    MBConv-1 & $16\sim32$ & $1\sim2$ & $1$ & $1$ \\
    MBConv-2 & $32\sim64$ & $2\sim6$ & $4\sim6$ & $2$ \\
    \hdashline[2pt/5pt]
    MBConv-3 & $32\sim64$ & $2\sim6$ & $4\sim6$ & $2$ \\
    ViT-3 & $24\sim64$ & $0\sim1$ & $1.0\sim2.0$ & - \\
    \hdashline[2pt/5pt]
    MBConv-4-1 & $64\sim96$ & $1\sim3$ & $4\sim6$ & $2$ \\
    ViT-4-1 & $48\sim96$ & $0\sim2$ & $1.0\sim2.0$ & - \\
    MBConv-4-2 & $64\sim96$ & $0\sim3$ & $4\sim6$ & $1$ \\
    ViT-4-2 & $48\sim96$ & $0\sim2$ & $1.0\sim2.0$ & - \\
    \hdashline[2pt/5pt]
    MBConv-5-1 & $96\sim128$ & $3\sim4$ & $4\sim6$ & $1$ \\
    ViT-5-1 & $64\sim128$ & $0\sim2$ & $1.0\sim2.0$ & - \\
    MBConv-5-2 & $96\sim128$ & $0\sim4$ & $4\sim6$ & $1$ \\
    ViT-5-2 & $64\sim128$ & $0\sim2$ & $1.0\sim2.0$ & - \\
    \hdashline[2pt/5pt]
    MBConv-6-1 & $192\sim224$ & $2\sim4$ & $4\sim6$ & $2$ \\
    ViT-6-1 & $144\sim224$ & $0\sim2$ & $1.0\sim2.0$ & - \\
    MBConv-6-2 & $192\sim224$ & $0\sim4$ & $4\sim6$ & $1$ \\
    ViT-6-2 & $144\sim224$ & $0\sim2$ & $1.0\sim2.0$ & - \\
    \hdashline[2pt/5pt]
    MBConv-7 & $224\sim240$ & $1\sim2$ & $6$ & $1$ \\
    ViT-7 & $176\sim240$ & $0\sim3$ & $1.0\sim2.0$ & - \\
    \hdashline[2pt/5pt]
    MBPool & $1792\sim1984$ & $1$ & $6$ & - \\
    \hline
    Input Resolution & \multicolumn{4}{c|}{$\{192,224,256,288\}$} \\
    \hline
    \end{tabular}
\caption{Our H4H-NAS search space. MBConv refers to IRB in Mobilenet-v2~\cite{sandler2018mobilenetv2}. ViT is from Efficientformer~\cite{li2022efficientformer}. MBPool is an efficient last stage~\cite{howard2019searching}.\vspace{-3em}}
\label{tab:searchs_space}
\end{table}

\myparagraph{Supernet Training}
We train our supernet for 360 epoches on the ImageNet dataset. 
At each training step, to optimize both supernet and subnets, we use the sandwich sampling rule~\cite{yu2019universally} to select four subnets: the smallest subnet, the largest subnet (i.e., the supernet), and two randomly sampled subnets in the search space.
As suggested by LeViT~\cite{graham2021levit} and NASViT~\cite{gong2022nasvit}, we use the AdamW optimizer when training our proposed search space. 
However, we use SGD for some CNN-only supernets used as baselines. 
We use an averaged gradient over the four sampled subnets during training, choose both dropout and drop-connect rate to be $0.2$, and use AutoAugment~\cite{cubuk2019autoaugment}. 

\myparagraph{Subnet Searching}
After supernet training is completed, we use evolutionary search 
to search for the optimal subnets given stringent system constraints on energy and latency.

\subsection{Performance Modeling and Scheduling}
\label{subsec:method_profiler}

We model heterogeneous AI edge devices consisting of both NPU and CIM macros. Our system model breaks down model inferences in fine granularity.
For convolution layers, it partitions the execution of different channels onto different devices. For transformer layers, the generation of Q/K/V and the execution of different heads in attention layers can be partitioned. The system modeling tool is used during evolutionary search in our H4H-NAS framework to profile the execution of a subnet.
The system modeling combines measurement results using a custom silicon and simulation results from industrial CIM IPs.
This modeling thus gives us an accurate latency and energy estimation for subnets generated in H4H-NAS.

\section{Evaluation Results}
\label{sec:evaluation}

In this section, we present results of co-designing in hybrid CNN/ViT models on the heterogeneous NPU-CIM system, followed by an ablation study in \Cref{sec:ablation}.

\begin{figure}[t]
    \centering
    \includegraphics[width=\linewidth]{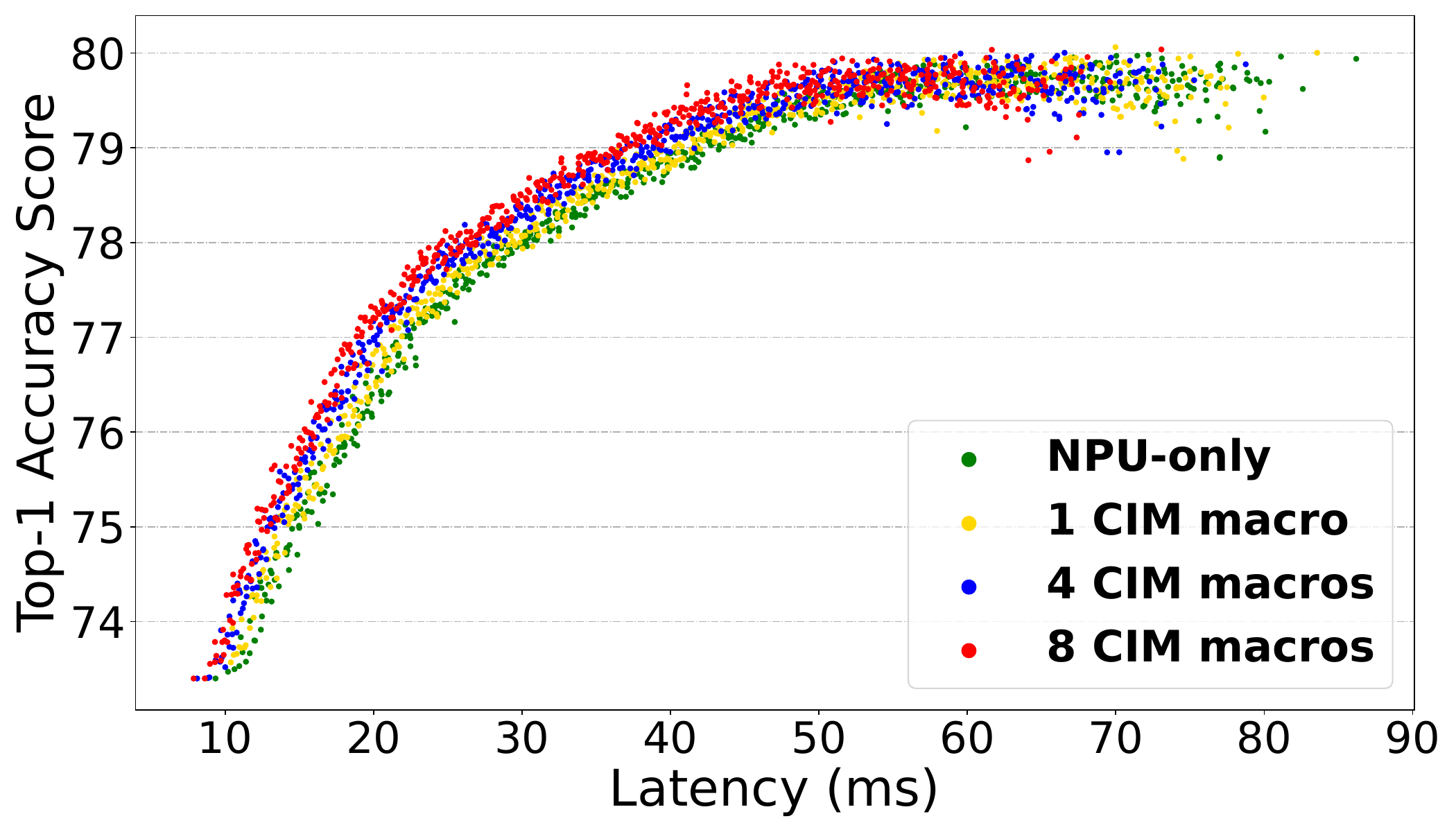}
    \vspace{-2.2em}
    \caption{H4H-NAS results searched from our hybrid CNN/ViT search space given latency constraints on different numbers of CIM macros.\vspace{-1em}}
\label{fig:vit_latency}
\end{figure}

\myparagraph{Heterogeneous Systems Reduce Hybrid Model Latency}
\Cref{fig:vit_latency} illustrates the results of latency-oriented search of our hybrid models in systems with different number of CIM macros.
It highlights that introducing heterogeneity into AI edge HW significantly reduces the inference latency. Meanwhile, given the same latency requirement, a system with 8 CIM macros can support a hybrid model with a $1.341\%$ higher top-1 accuracy, compared with a NPU-only system. In addition, when acquiring models with the same accuracy, a system with 8 CIM macros can perform the inference with an average $21.99\%$ and up to $56.08\%$ latency reduction.

\begin{figure}[t]
    \centering
    \includegraphics[width=\linewidth]{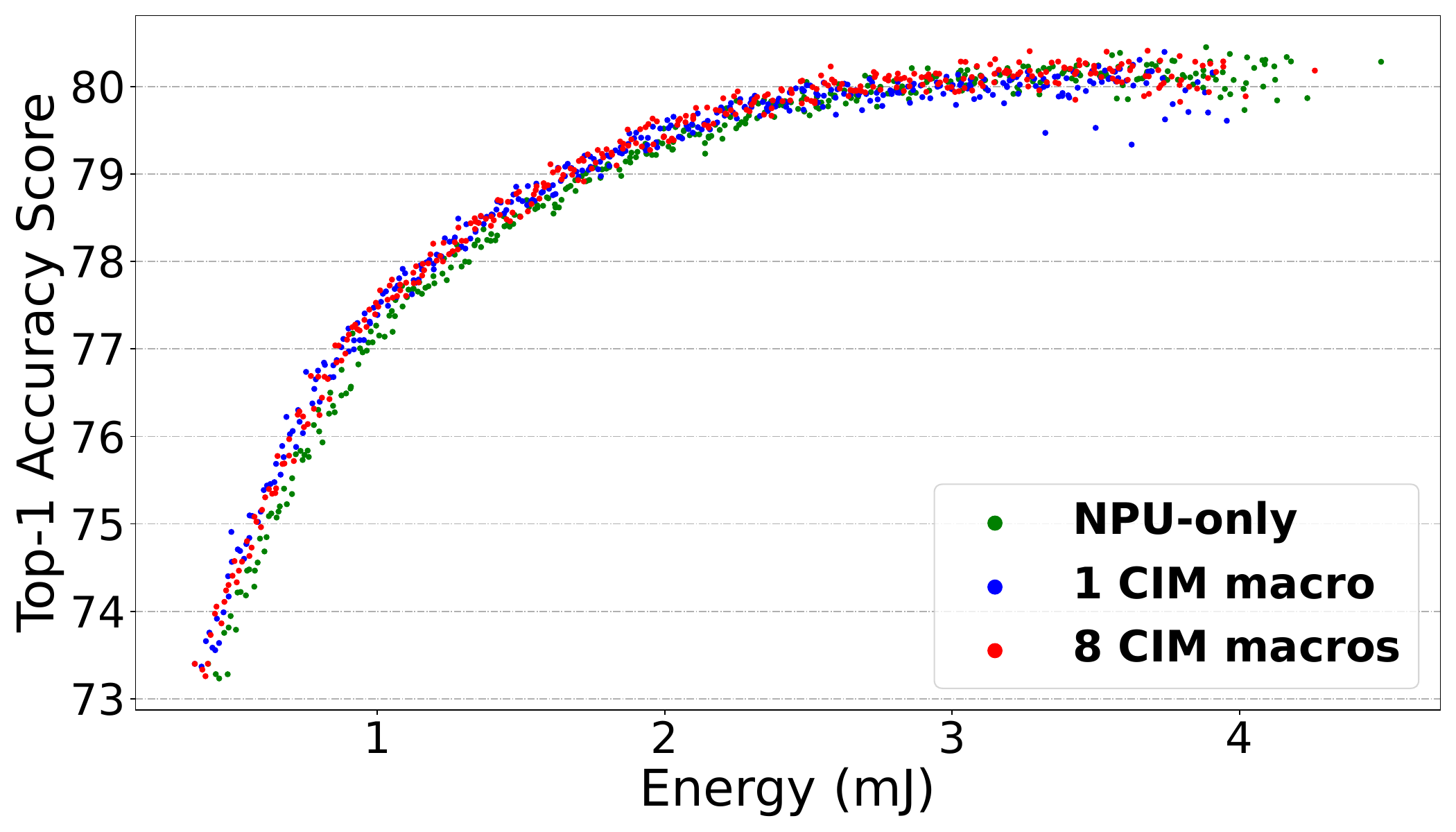}
    \vspace{-2.5em}
    \caption{H4H-NAS results from hybrid CNN/ViT given energy constraints on different numbers of CIM macros.\vspace{-1em}}
\label{fig:vit_energy}
\end{figure}

\myparagraph{Heterogeneous Systems Save Energy}
In \Cref{fig:vit_energy}, we show the results of energy-oriented search of our hybrid models in systems with different numbers of CIM macros. Similar to latency, heterogeneous systems improve energy efficiency. Given the same energy requirement, a system with 8 CIM macros can support a hybrid model with $0.614\%$ higher accuracy. Meanwhile, fixing accuracy, a system with 8 CIM macros performs inference with an average $11.80\%$ less energy consumption and up to $33.13\%$ energy reduction.

\myparagraph{Effects of Multiple CIM Macros}
Compute can be parallelized on CIMs when more than one CIM is available in the system. As shown in \Cref{fig:vit_latency}, more CIMs reduce overall inference latency.
However, the improvement in energy efficiency does not scale with more CIMs introduced, as shown in \Cref{fig:vit_energy}.
This is because, intuitively, adding more macros without changing its internal structure does not significantly change the unit energy-efficiency of operations.

\section{Ablation Study}
\label{sec:ablation}

\subsection{ResNet vs.~MobileNet-v2 vs.~Hybrid Model}
\begin{figure}[t]
    \centering
    \includegraphics[width=0.9\linewidth]{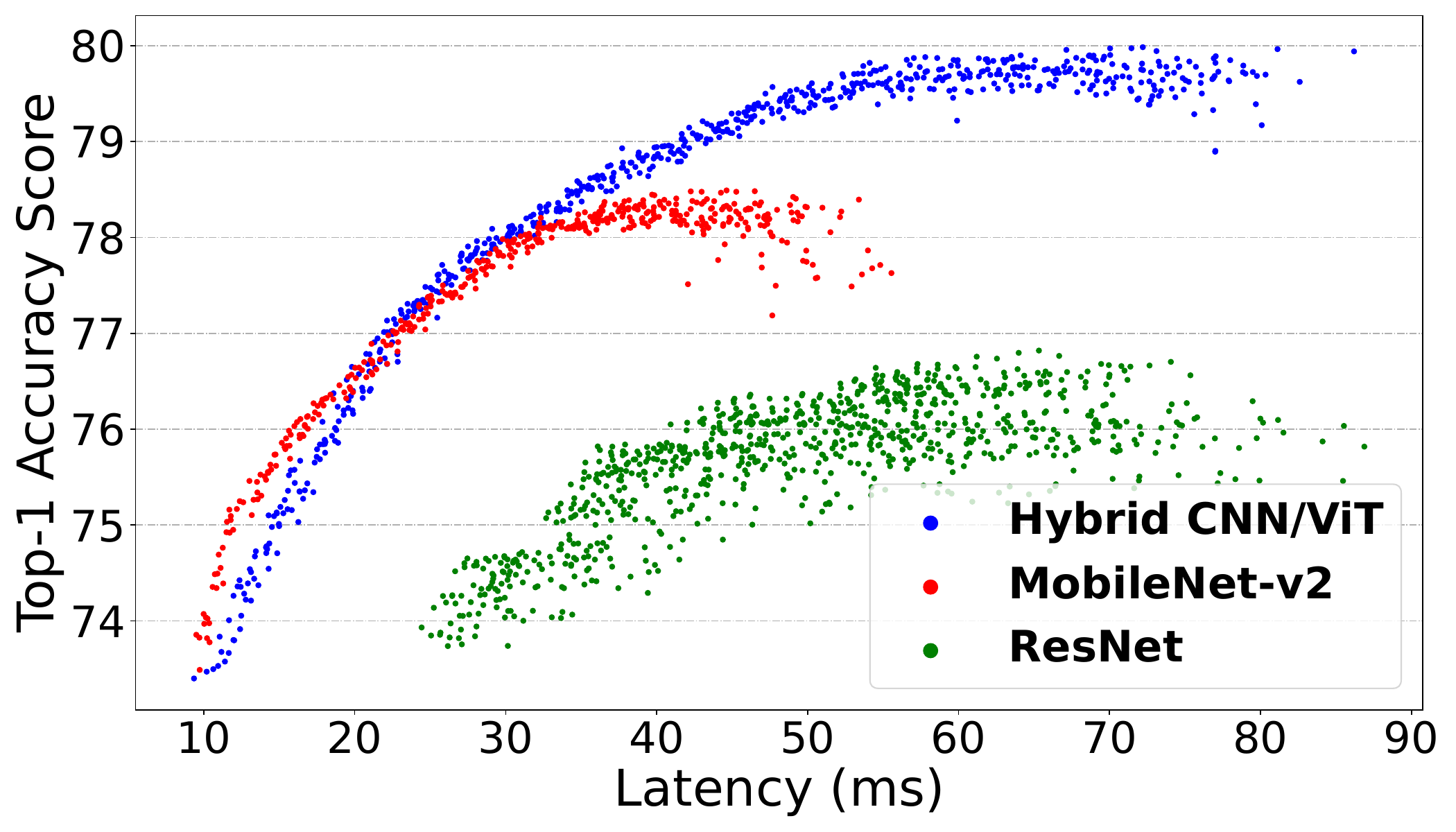}
    \vspace{-1.2em}
    \caption{H4H-NAS results of Hybrid CNN/ViT vs.~MobileNet-v2 vs.~ResNet, given latency constraints on NPU-only systems.\vspace{-1em}}
\label{fig:different_model_types}
\end{figure}

To evaluate the efficacy of hybrid model architectures for heterogeneous AI edge systems, we first conduct searches for the optimal models based on ResNet, MobileNet-v2 (IRB) and hybrid CNN/ViT structures, when executing on an NPU-only system.
As shown in~\Cref{fig:different_model_types}, the top-1 accuracy of subnets increases as more latency is afforded. For medium-sized models ($>$20ms), hybrid models achieve significantly better performance than IRB-based models given the same latency/energy budget.
Also note that IRB-based models achieve good accuracy/latency trade-off under extreme latency budgets ($<$20ms) and strictly outperform ResNet counterparts given all constraints.
This also aligns with most AI edge applications using IRB instead of residue blocks for better performance.

\begin{figure}[t]
    \centering
    \includegraphics[width=0.85\linewidth]{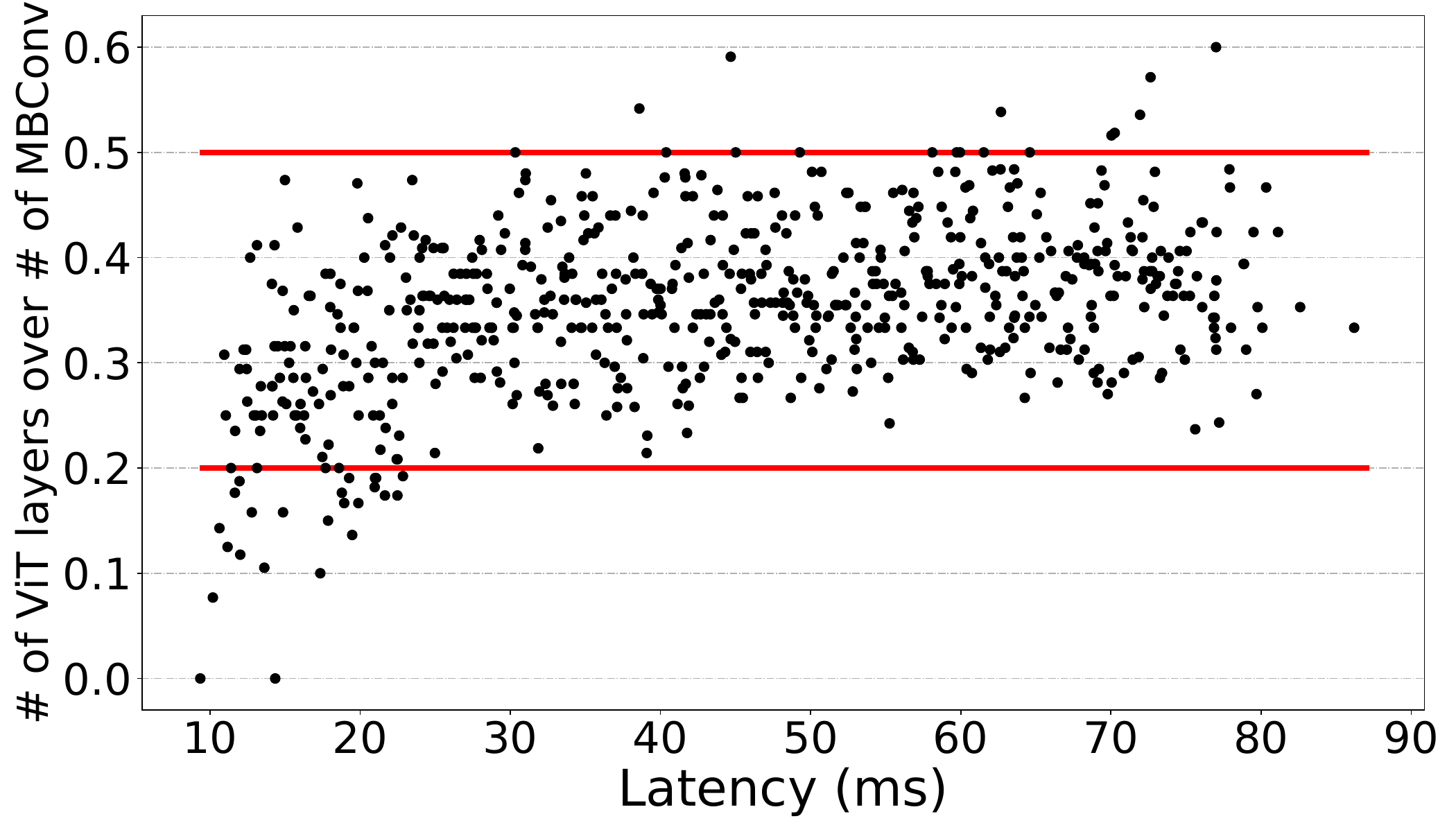}
    \vspace{-1.5em}
    \caption{Comparative ratio between number of MBConv layers over number of ViT layers, in each subnet.\vspace{-1.5em}}
\label{fig:vit_ratio}
\end{figure}

\myparagraph{Efficient CNN/ViT Basic Block}
As another by-product, \Cref{fig:vit_ratio} shows the ratio of number of ViT blocks over the number of IRB blocks, given different latency constraints.
Interestingly, H4H-NAS tends to include both CNN and ViT blocks and keep a balance between IRB/ViT blocks for efficient edge implementation.
All searched subnets have a similar proportion of 2--5 ViT + $10$ IRB (between the red lines of \Cref{fig:vit_ratio}).

The finding aligns with recent hand-crafted hybrid architectures~\cite{park2021vision} as IRB and ViT are complementary when executing CV tasks.
IRB abstracts neighbouring information in a feature map into tokens and ViT then translates the token embedding using attention layers.
Hybrid architectures thus provide better accuracy/performance trade-offs.

\subsection{Increased Parallelism inside a CIM Macro}

\begin{figure}[t]
    \centering
    \includegraphics[width=\linewidth]{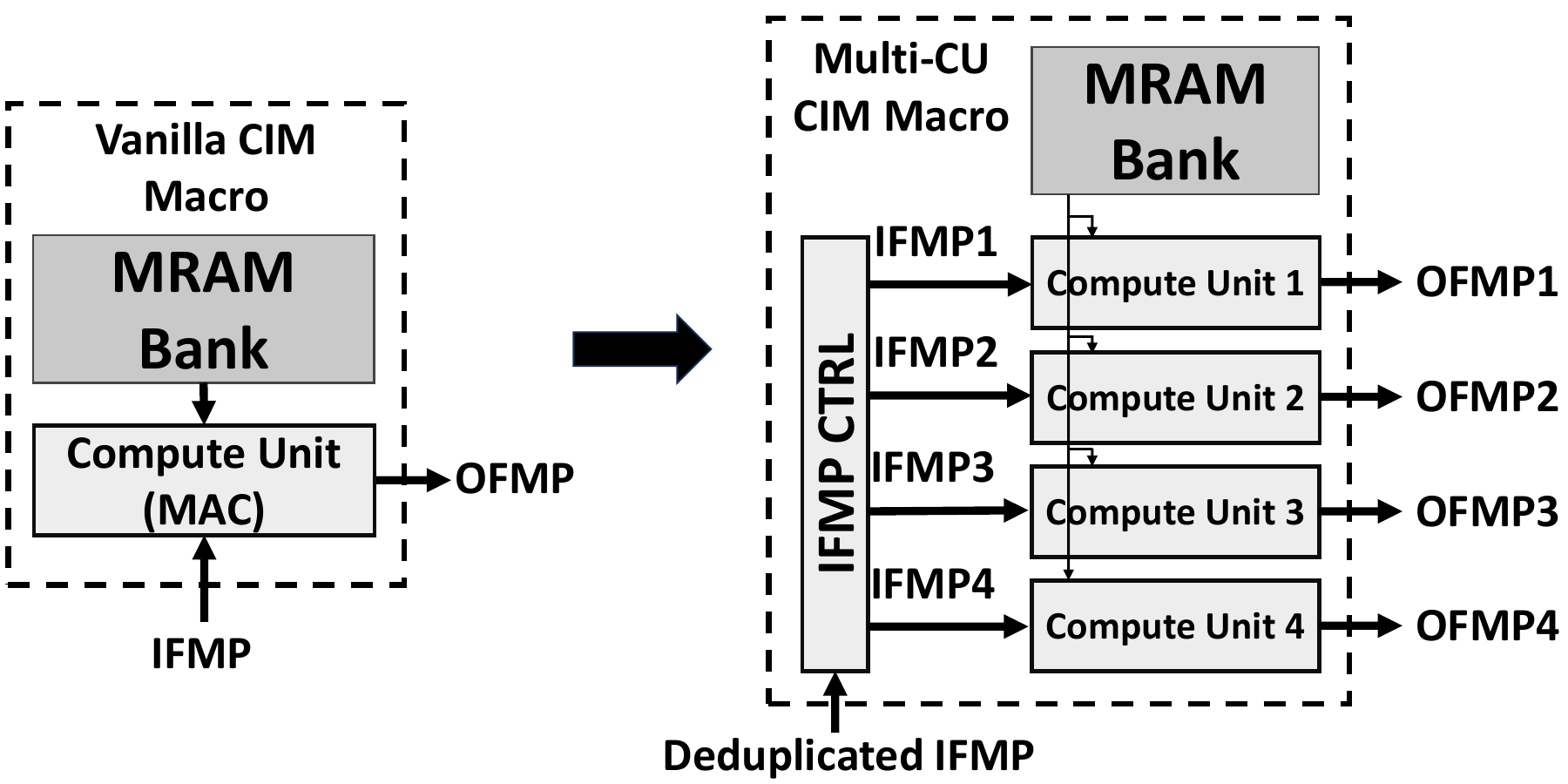}
    \vspace{-2em}
    \caption{An example of introducing four compute units into one CIM macro.\vspace{-1.5em}}
\label{fig:multi_cu_cim}
\end{figure}

In \Cref{sec:evaluation}, we showed that using multiple CIM-macros improves inference latency over a \sota single-macro.
Here, we explore the improvement of introducing multiple compute units inside one single CIM macro. \Cref{fig:multi_cu_cim} shows an example of a single CIM macro that has four compute units inside.

This design is promising in two aspects.
Firstly, it provides another level of parallelism in computation.
Secondly, some repeated input data can be merged and transferred together into the CIM macro. The input feature map (IFMP) controller reorganizes the dataflow required for computation.
For instance, depthwise convolutions can benefit from input deduplication if adjacent output elements are computed simultaneously.
Its theoretical read reduction can reach $2/3$ with (3,3)-kernel, (1,1)-stride and large inputs.
\Cref{tab:area_data} shows the area results of designing multiple compute-units into one macro, indicating that such designs will only cause acceptable overheads in area sizes.

\begin{table}
\centering
    \begin{tabular}{|c|c|c|}
    \hline
    Bit-cell Area & Area Overhead per CU & IFMP CTRL Area \\
    \hline
    $0.0063\mu m^2$ & 14$\%$ & $<$0.1$\%$ \\
    \hline
    \end{tabular}
\caption{Bit-cell area of MRAM projected in 7nm; Area overhead each time when introducing a new compute-unit; Synthesized area overhead of the IFMP controller.\vspace{-2.5em}}
\label{tab:area_data}
\end{table}

\begin{figure}[t]
    \centering
    \includegraphics[width=\linewidth]{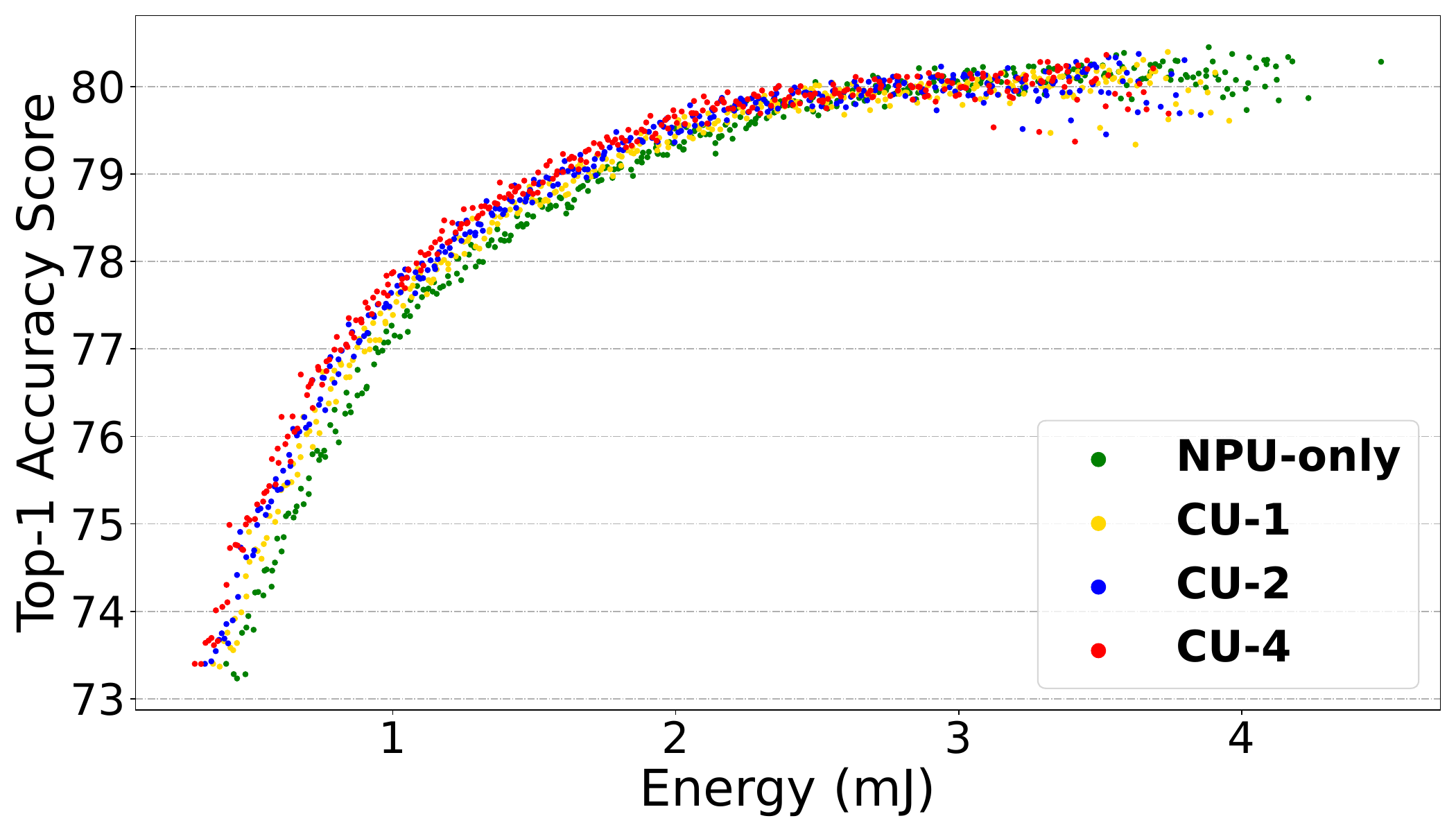}
    \vspace{-2.5em}
    \caption{Energy-oriented H4H-NAS for a single-macro system with different numbers of compute units.\vspace{-1.5em}}
\label{fig:multi_compute_units}
\end{figure}

We integrate and evaluate the multi-CU design into our NAS framework.
\Cref{fig:multi_compute_units} shows energy-oriented searches on single-macro systems with different numbers of compute units.
Given the same top-1 accuracy, a single-macro system with 4 compute units reduces an average $19.11\%$ and up to 41.72$\%$ energy consumption from an NPU-only system, and an average $9.34\%$ from a vanilla NPU+CIM system with one compute unit per macro.

\section{Conclusion}
This paper presents H4H-NAS, a NAS framework to design efficient hybrid CNN/ViT models for heterogeneous edge systems with both NPU and CIM.
The framework provides up to $1.34\%$ top-1 accuracy improvement, and up to 56.08$\%$ latency and 41.72$\%$ energy improvements.
Key techniques include a highly-flexible hybrid model search space, a reliable performance profiler for heterogeneous systems, and system improvements with increased CIM parallelism.
Our framework is adaptable to future edge device designs, and conversely sheds light on how to design ML models and edge systems.

\renewcommand\acksname{\textsc{ACKNOWLEDGMENTS}}
\begin{acks}
This research was supported by Meta Reality Labs Research, TSMC Corporate Research, NSF grants CCF-1919223 and CNS-2211882, Parallel Data Lab (PDL) Consortium, and the Lee-Stanziale Ohana Endowed Fellowship.
\end{acks}

\renewcommand\refname{\textsc{REFERENCES}}
\bibliographystyle{ACM-Reference-Format}
\bibliography{reference}


\end{document}